\documentclass[a4paper,fleqn]{cas-dc}

\usepackage[numbers]{natbib}

\usepackage{subcaption}

\def\tsc#1{\csdef{#1}{\textsc{\lowercase{#1}}\xspace}}
\tsc{WGM}
\tsc{QE}
\tsc{EP}
\tsc{PMS}
\tsc{BEC}
\tsc{DE}
\begin{document}
\let\WriteBookmarks\relax
\def\floatpagepagefraction{1}
\def\textpagefraction{.001}
\shorttitle{ExAID: A Multimodal XAI Framework for CAD of Skin Lesions}
\shortauthors{Lucieri~et~al.}

\title [mode = title]{ExAID: A Multimodal Explanation Framework\\for Computer-Aided Diagnosis of Skin Lesions}                      

\author[1,2]{Adriano Lucieri}[]
\cormark[1]
\ead{adriano.lucieri@dfki.de}

\author[1,2]{Muhammad Naseer Bajwa}[]
\ead{naseer.bajwa@dfki.de}
\author[3,4]{Stephan Alexander Braun}[]
\ead{stephanalexander.braun@ukmuenster.de}
\author[5,6]{Muhammad Imran Malik}[]
\ead{malik.imran@seecs.edu.pk}
\author[1,2]{Andreas Dengel}[]
\ead{andreas.dengel@dfki.de}
\author[1]{Sheraz Ahmed}[]
\ead{sheraz.ahmed@dfki.de}

\address[1]{German Research Center for Artificial Intelligence (DFKI) GmbH, Trippstadter Strasse 122, 67663 Kaiserslautern, Germany}

\address[2]{Technical University Kaiserslautern, Erwin-Schrödinger-Strasse 52, 67663 Kaiserslautern, Germany}

\address[3]{University Hospital M\"unster, 48149 M\"unster, Germany}

\address[4]{University Hospital of D\"usseldorf, 40225 D\"usseldorf, Germany}

\address[5]{School of Electrical Engineering and Computer Science (SEECS), National University of Sciences and Technology (NUST), Islamabad, Pakistan}

\address[6]{Deep Learning Laboratory, National Center of Artificial Intelligence, Islamabad, Pakistan}

\cortext[cor1]{Corresponding author}

\begin{abstract}
\phantom\textbf{Background and Objectives:} One principal impediment in successful deployment of Artificial Intelligence (AI)-based Computer-Aided Diagnosis (CAD) systems in everyday clinical workflow is their lack of transparent decision making. Although commonly used eXplainable AI (XAI) methods provide some insight into these largely opaque algorithms, yet such explanations are usually convoluted and not readily comprehensible except by highly trained AI experts. The explanation of decisions regarding the malignancy of skin lesions from dermoscopic images demands particular clarity, as the underlying medical problem definition is itself ambiguous. This work presents and evaluates ExAID (Explainable AI for Dermatology), a novel XAI framework for biomedical image analysis, providing multi-modal concept-based explanations consisting of easy-to-understand textual explanations supplemented by visual maps justifying the predictions.

\noindent\textbf{Methods:} Our framework relies on Concept Activation Vectors (CAVs) to map human-understandable concepts to those learnt by an arbitrary Deep Learning based algorithm in its latent space, and Concept Localisation Maps (CLMs) to highlight concepts in the input space. This identification of relevant concepts is then used to construct fine-grained textual explanations supplemented by concept-wise location information to provide comprehensive and coherent multi-modal explanations. All decision-related information is comprehensively presented in a diagnostic interface for use in clinical routines. Moreover, the framework includes an educational mode providing dataset-level explanation statistics and tools for data and model exploration to aid medical research and education processes.

\noindent\textbf{Results:} Through rigorous quantitative and qualitative evaluation of our framework on a range of dermoscopic image datasets such as SkinL2, Derm7pt, PH\textsuperscript{2} and ISIC, we show the utility of multi-modal explanations for CAD-assisted scenarios even in case of wrong disease predictions.

\noindent\textbf{Conclusions:} We present a new multi-modal explanation framework for biomedical image analysis on the example use-case of Melanoma classification from dermoscopic images and evaluate its utility on a row of datasets. Since comprehensible explanation is one of the cornerstones of any CAD system, we believe that ExAID will provide dermatologists an effective screening tool that they both understand and trust. Moreover, ExAID will be the basis for similar applications in other biomedical imaging fields.

\end{abstract}


\ExplSyntaxOn
\keys_set:nn { stm / mktitle } { nologo }
\ExplSyntaxOff
\maketitle

\section{Introduction}
\label{sec:introduction}

In 2016, Ribeiro et al.~\cite{ribeiro2016should} reported an image classifier that was able to inadvertently classify correctly but for wrong reasons. 
They found out that their wolf versus dog classifier learnt an undesirable correlation between wolf and background snow and, therefore, would classify a given image as wolf if there was snow in the background. 
If it were not due to authors' vigilance in finding explanations to the model's predictions, it would have been difficult to properly evaluate trustworthiness of this image classifier. 
Although this was an inconsequential example of spurious correlations learnt from a large amount of data, wrong decisions in safety critical domains resulting from such misunderstandings can potentially have grave impact on human lives.
The application of Artificial Intelligence (AI) methods on medical tasks has become ubiquitous in the last decade~\cite{al2020evaluation, khan2020coronet, sunija2021octnet}.
Therefore, hesitation of medical practitioners in trusting diagnostic predictions of any automated Computer-Aided Diagnosis (CAD) system is understandable since such systems normally provide little to no cognisance regarding their decision making process~\cite{lucieri2020achievements}. 
In addition to evaluating the reasons behind a model's predictions, explanation methods can also help in revealing new diagnostic criteria~\cite{izadyyazdanabadi2018weakly} previously unknown to medical practitioners.
The requirement for a CAD to be explainable arose with early applications of AI in healthcare and became more relevant with recent ethical and legal standards~\cite{arrieta2020explainable,GDPR2018}. 
The consequent increase in research activity in the domain of eXplainable AI (XAI) also reflects the growing interest of the community to provide explanations for CAD systems~\cite{arrieta2020explainable}.

XAI methods for the image-domain come in a variety of forms and provide explanations using a range of modalities such as feature-relevance visualisations~\cite{ribeiro2016should, fong2019understanding, wang2020score, chen2019looks}, textual explanations~\cite{hendricks2018grounding, zhang2017mdnet}, or quantitative relevance measures for abstract concepts~\cite{kim2018CAV, chen2020concept}. 
They differ not only in the way they are presented to their users but also in their derivation, resulting in varying levels of insight provided regarding the decision making of the AI.
Furthermore, these methods can be either ante-hoc (e.g. ProtoPNet~\cite{chen2019looks}, MDNet~\cite{zhang2017mdnet}), with a decision making process that is explainable by design, or post-hoc, providing explanations for an AI model after construction and training of the model using model-specific (e.g. Score-CAM~\cite{wang2020score}, TCAV~\cite{kim2018CAV}) or model-agnostic (e.g. LIME~\cite{ribeiro2016should}, EP~\cite{fong2019understanding}) techniques~\cite{murdoch2019definitions}.
Most methods provide explanations on a local scale (individual data samples) while some aim at approximating explanations on a global scale (holistic model behaviour). 
However, model explanations given by single XAI methods are usually not sufficient to provide plausible and easy-to-understand decision justification to end users.

Melanoma is the most dangerous skin cancer, leading to the majority of skin-related deaths in the US while accounting for only 1\% of skin cancers diagnosed~\cite{american20202cancer}. 
Regular preventive examinations are conducted by physicians through naked-eye observation or dermoscopic imaging.
In dermoscopic pattern recognition, experts look for dermoscopic criteria and apply manual algorithms like the ABCD-rule~\cite{nachbar1994abcd} or 7-point checklist~\cite{argenziano1998epiluminescence} to judge the malignancy of a lesion. 
Currently, AI-based dermatology focuses mostly on the analysis of dermoscopic images~\cite{maglogiannis2015enhancing, tschandl2019expert, mahbod2020effects}.
However, first approaches towards analysing raw, clinical images have been proposed as well~\cite{BIRKENFELD2020105631}.
The majority of explanation approaches for dermoscopic skin lesion analysis rely on the application of visual XAI through saliency maps~\cite{xiang2019towards, young2019deep} or attention mechanisms~\cite{gu2020net, barata2021explainable}. 
Another common approach is the detection and localisation of dermoscopic criteria as used by doctors in manual classification.
Coppola et al.~\cite{coppola2020interpreting}, for instance, train a multi-task CNN predicting dermoscopic features with information sharing between different subnetworks to increase interpretability.
In~\cite{lucieri2020interpretability}, Lucieri et al. apply the concept-based TCAV method to predict dermoscopic criteria from a pre-trained network to explain its predictions.
Dermoscocpic criteria localisation has been been approached by combining TCAV with perturbation-based saliency methods in~\cite{lucieri2020CLM}
and through explicit segmentation of criteria in \cite{sonntag2020skincare}.
For a complete survey on XAI in Dermatology the reader is referred to~\cite{lucieri2021deep}.

Several frameworks for AI-based medical imaging have been proposed in recent years~\cite{gibson2018niftynet, sonntag2020skincare, hasan2021dermo, jiang2021visually}.
While some lack proper and comprehensible explainability, others do not provide an easy-to-use interface for human-machine interaction, impeding the utilisation in diagnostic routines or research.
Moreover, first commercial platforms for biomedical AI exist~\cite{tosun2020histomapr, datalanguage2021, decodedhealth2021, hacarus2021}, claiming to provide explanations for their algorithms.

In this paper we present a novel XAI framework, namely Explainable Artificial Intelligence for Dermatology (ExAID)\footnote{A demo will be soon available under https://exaid.kl.dfki.de/.} which is able to provide easy-to-understand textual, visual and conceptual explanations for automated analysis of dermoscopic images of malignant and benign skin lesions, while being adaptable to any other biomedical imaging use case. 
ExAID is built upon two of our previous works: \cite{lucieri2020interpretability}, which verifies that deep learning models are able to learn and utilise similar disease-related concepts as described by dermatologists and employed by them during manual analysis of skin images; and \cite{lucieri2020CLM}, which localises these concepts, learnt and embedded in the latent space of the model, on the original image. 
ExAID extends the previously proposed explanation modalities by introducing concept-based textual explanations while integrating all modalities in a unified, intuitive framework to further enhance intelligibility of AI's decision making in a diagnostic setting.
In addition to clinical diagnosis functionalities, it provides in-depth analysis tools for medical research and education. 
Therefore, the framework offers two distinct interfaces for clinical diagnosis and research purposes, laying the foundation for understandable and transparent integration of AI in medical workflows.
In contrast to existing XAI frameworks, ExAID emphasises intuitive intelligibility for end users by conveying multi-modal decision justification centred around standard concepts commonly used in the dermatology domain.

The rest of the paper is organised as follows. 

Section~\ref{sec:materials_and_methods} covers details of the datasets used for training and evaluation as well as an introduction to the framework and its components.
Experimental setups for the generation of explanations and their evaluation are described in Section~\ref{sec:experiments_and_results}.
Finally, Section~\ref{sec:discussion} discusses the presented results and elaborates on current limitations of this particular use case and framework state, before the work is concluded in Section~\ref{sec:conclusion}. 

%%%%%%%%%%%%%%%%%%%%%%%%%%%%%%%%%%%%% Materials and Method %%%%%%%%%%%%%%%%%%%%%%%%%%%%%%%%%%%%%

\section{Methods}
\label{sec:materials_and_methods}

\subsection{Datasets}
ExAID contains two types of classifiers: Disease-level classifiers for lesion diagnosis and concept-level classifiers for detection of dermatological concepts in a given image. 
To train these two classifiers, it requires datasets with two types of labels, namely disease labels (\textit{Melanoma} and \textit{Nevus}) and concept annotations (presence or absence of dermoscopic concepts).

\subsubsection{Datasets for Disease-level Classification}
The training set for disease-level classification consists of \textit{Melanoma} and \textit{Nevi} images taken from ISIC 2019 as well as PH\textsuperscript{2}~\cite{mendoncca2013ph} and derm7pt~\cite{Kawahara2018-7pt} datasets. 
ISIC 2019 dataset is a public collection of 25,331 images of different provenance divided into eight different classes. 
This dataset is a coalition of three datasets, HAM10000~\cite{tschandl2018ham10000}, BCN20000~\cite{combalia2019bcn20000}, and MSK~\cite{codella2018MSK}. 
Since the common denominator of ISIC2019, PH\textsuperscript{2}, and derm7pt datasets are \textit{Melanoma} and \textit{Nevi} classes, we assembled a subset of the three datasets consisting of images from these two classes only, and manually cleansed the dataset for duplicates and samples with low quality (e.g. systematic artefacts), resulting in a total of 6,475 images. 
As PH\textsuperscript{2} and derm7pt will be used for training the concept-level classifiers, a custom dataset split is assembled from a combination of all three stratified datasets to avoid covariate shifts between disease-level and concept-level training stages.
The distribution of images in training, validation and test sets for disease-level classification is given in Table~\ref{tab:dataDistImgClf}. 
The generalisability of the model is moreover evaluated on a range of other datasets including 2016 and 2017 ISIC challenge datasets and SKINL2~\cite{de2019light} dataset as shown in Table~\ref{tab:resultsImgClf}.

%%%%%%%%%%%%%%%%%%%%%%%%%%%%%%%% Data Distribution for Disease Classification %%%%%%%%%%%%%%%%%%%%%%
\begin{table}[t!]
\centering
\caption{Distribution of data in training, validation and test splits for disease-level classification.}
\label{tab:dataDistImgClf}
\begin{tabular}{llccc}
\hline
\multicolumn{1}{l}{\multirow{2}{*}{\textbf{Split}}} & \multicolumn{1}{c}{\multirow{2}{*}{\textbf{Dataset}}} & \multicolumn{3}{c}{\textbf{Lesions}}                \\ \cline{3-5} 
\multicolumn{1}{c}{}                                & \multicolumn{1}{c}{}                                  & \textbf{Melanoma} & \textbf{Nevi} & \textbf{Total} \\ \hline
\multirow{3}{*}{\textbf{Train}}    & ISIC2019 & 1250 & 2894 & 4144 \\
                                   & Derm7pt  & 158 & 368 & 526  \\
                                   & PH2      & 26 & 102 & 128  \\ \hline
\multirow{3}{*}{\textbf{Validate}} & ISIC2019 & 313 & 723 & 1036 \\
                                   & Derm7pt  & 40 & 92 & 132  \\
                                   & PH2      & 6 & 26 & 32   \\ \hline
\multirow{3}{*}{\textbf{Test}}     & ISIC2019 & 391 & 904 & 1295 \\
                                   & Derm7pt  & 50 & 115 & 165  \\
                                   & PH2      & 8 & 32 & 40   \\ \hline
\multirow{3}{*}{\textbf{Total}}    & ISIC2019 & 1954 & 4521 & 6475 \\
                                   & Derm7pt  & 248 & 575 & 823  \\
                                   & PH2      & 40 & 160 & 200  \\ \hline
\end{tabular}
\end{table}

%%%%%%%%%%%%%%%%%%%%%%%%%%%%%%%% Data Distribution for Concept Classification %%%%%%%%%%%%%%%%%%%%%%
\begin{table}[b!]
\centering
\caption{Distribution of data in training, validation and test splits for concept-level classification with D7PH2 dataset.}
\label{tab:dataDistConceptClf}
\begin{tabular}{llccc}
\hline
\multicolumn{1}{l}{\multirow{2}{*}{\textbf{Split}}} & \multicolumn{1}{c}{\multirow{2}{*}{\textbf{Dataset}}} & \multicolumn{3}{c}{\textbf{Lesions}}                \\ \cline{3-5} 
\multicolumn{1}{c}{}                                & \multicolumn{1}{c}{}                                  & \textbf{Melanoma} & \textbf{Nevi} & \textbf{Total} \\ \hline
\multirow{2}{*}{\textbf{Train}}    & Derm7pt & 158 & 368 & 526 \\
                                   & PH2     & 26 & 102 & 128 \\ \hline
\multirow{2}{*}{\textbf{Validate}} & Derm7pt & 40 & 92 & 132 \\
                                   & PH2     & 6 & 26 & 32 \\ \hline
\multirow{2}{*}{\textbf{Test}}     & Derm7pt & 50 & 115 & 165 \\
                                   & PH2     & 8 & 32 & 40 \\ \hline
\multirow{2}{*}{\textbf{Total}}    & Derm7pt & 248 & 575 & 823 \\
                                   & PH2     & 40 & 160 & 200 \\ \hline
\end{tabular}
\end{table}

%%%%%%%%%%%%%%%%%%%%%%%%%%%%%%%% ExAID - Schematic Drawing %%%%%%%%%%%%%%%%%%%%%%
\begin{figure*} % ExAID Figure
    \centering
    \includegraphics[width = 0.9\textwidth]{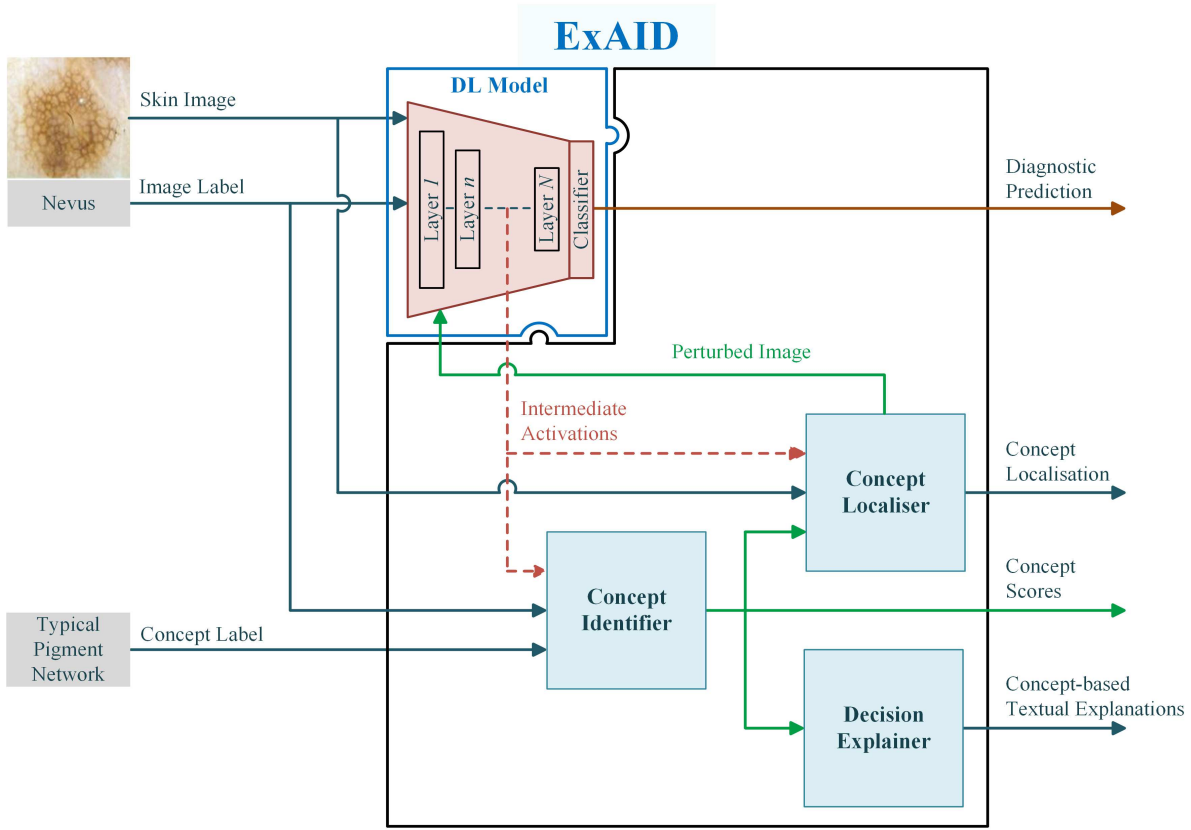}
    \caption{ExAID Framework architecture. The schematic drawing shows the in-, output and flow of information through ExAID as well as the relationship between its components.}
    \label{fig:exaid}
\end{figure*}

\subsubsection{Datasets for Concept-level Classification}

Training of concept classifiers requires annotations regarding presence or absence of specific dermoscopic concepts.
These annotations are not usually available with dermoscopic image datasets, which limits our selection of training and evaluation datasets primarily to PH2 and derm7pt. 
The PH\textsuperscript{2} dataset is a small dataset of only 200 dermoscopic images containing 80 common nevi, 80 atypical nevi, and 40 melanoma. 
For each image, the dataset provides colour and lesion segmentation masks and extensive, well-curated annotations with respect to presence or absence of various concepts. 
The derm7pt dataset contains 1,011 clinical and dermoscopic images divided into four diagnosis classes and one miscellaneous class. 
Two of these diagnosis classes, \textit{Melanoma} and \textit{Nevi} are further divided into 13 sub-classes. 
From this dataset, 823 images belonging to only \textit{Melanoma} and \textit{Nevi} samples have been considered. 
The combination of derm7pt and PH\textsuperscript{2} used for concept classification is subsequently referred to as D7PH2.
Table~\ref{tab:dataDistConceptClf} shows the distribution of images used in the concept-level classification task.
ISIC 2016 and 2017 challenge datasets are moreover used for the evaluation of concept classifier generalisability.
However, both datasets only include annotations of two dermoscopic concepts each, namely \textit{Pigment Networks} and \textit{Streaks} as well as \textit{Dots \& Globules} and \textit{Streaks}, respectively.

\subsection{ExAID Framework}

At its core, ExAID is a generic toolbox for human-centred post-hoc explanations able to explain arbitrary DL-based models even beyond applications in dermatology. 
In addition to the DL model to be explained, its computational foundation consists of three basic components, namely Concept Identification, Concept Localisation and Decision Explanation modules as depicted in Fig.~\ref{fig:exaid}. 

\subsubsection{Concept Identifier}
The Concept Identifier maps disease-related dermatological concepts to their corresponding representation learnt by the DL-based model in its latent space using Concept Activation Vectors~\cite{kim2018CAV} (CAVs).
For each pre-defined concept a linear binary classifier is trained on the detection of said concept from the model's activation space, resulting in a CAV which represents the main concept direction in this latent space.
CAV training can be executed on arbitrary model layers, automatically selecting each concept's best performing activations for inference.

Once trained, the concept classifiers allow to predict presence or absence of single concepts on unseen images, based on the model's latent activation patterns.
CAVs additionally allow for computation of the global TCAV metric, estimating a concept's overall contribution to the prediction of a certain target class. 
Further details on the Concept Identifier module and the CAV training procedure can be found in~\cite{lucieri2020interpretability}.

\subsubsection{Concept Localiser}
Concept Localisation Maps (CLMs)~\cite{lucieri2020CLM} extend CAVs by localising regions pertinent to a learned concept in the latent space of a trained image classifier. 
They provide qualitative and quantitative assurance of the model's ability to learn the right interpretation of a concept by indicating the exact spatial location that contributed to a concept prediction and moreover enable the visualisation of other, potentially abstract concepts.

Given an input image \( x \in X \), a linear concept classifier \( g_C \) generates a concept score for a concept \( C \) based on the trained model's latent vector \( f_l(x; \theta) \) at layer \( l \) with optimal weights \( \theta \).
The Concept Localiser implements the perturbation-based concept localisation technique from~\cite{lucieri2020CLM} to generate spatial importance values based on variation of the concept scores \( g_C(f_l(x; \theta)) \). 
The resulting map \( m_{Cl} \) corresponds to the input region contributing most to concept~$C$. 
Instead of occlusion by means of black patches, in this work a radial Gaussian mask is applied to a blurred image patch to mitigate distribution shift in perturbed images stemming from sharp edges and colour gradients in the perturbed images. 

\subsubsection{Decision Explainer}
The Decision Explainer receives all concept prediction scores for a given image from the Concept Identifier. 
A rule-base is derived from a calibration dataset and applied to the translation of single concept scores into a textual decision explanation grounded in human-understandable conceptual evidence. 
An explanation sentence conveys graded information about the conceptual evidence detected by a given model, as well as its influence to the given prediction.
An example for a textual explanation along with the corresponding input image is given in Fig.~\ref{fig:modes_diagnostic}.

The explanations derived from concept detection are composed into coherent and easy-to-understand explanation texts. 
An explanation sentence is constructed based on concept scores and directional derivatives computed during concept detection under discrimination between absence, moderate evidence and strong evidence of concepts to reflect the fuzzy nature of concepts' appearance. 
Manifestation of a concept is decided by means of thresholds derived from the concept training data. 
This is achieved by first scaling the unbound concept prediction using a two-sided normalisation scheme to obtain a centred probability of concept presence. 
Thresholds are then derived by maximising False Positive Rate (FPR) and True Positive Rate (TPR) among all positive predictions on the training dataset for moderate and strong evidence thresholds, respectively.
The directional derivatives of the predicted class along the individual CAV is used to indicate positive or negative influence of concept to the prediction.
Conceptual evidence is listed after the key word ``despite'' in case of negative class influence to signalise contraindication (see Figure~\ref{fig:examples_verbal}).

\subsection{Operation Modes}
ExAID offers two complementary operation modes that are meant for different use cases. 
A diagnostic mode provides functionality meant to support dermatologists during clinical examination of patient's skin lesions. 
For research and education purposes, ExAID offers an educational mode including a collection of tools for holistic analysis of the deep model's behaviour as well as the collected case data.

\subsubsection{Diagnostic Mode}
The majority of a dermatologist's clinical routine consists of visual examination of patients' skin lesions to reach a decision regarding further investigation of a potentially malignant lesion. 
Provided enough evidence for malignancy, the suspicious tissue may be excised under local anaesthesia. 
Physicians with considerable experience in dermoscopy develop an intuition which enables them to promptly reach to a conclusion while novices initially need to pay greater attention to the assessment of a particular skin lesion.
This is among other things owed to the disarray of dermoscopic terms and concepts and their usage in different schools of thought.
Having developed a routine and diagnostic intuition not only bears the risk of subjective bias in a decision, but it might also lead to negligence in the identification of important diagnostic details, which is furthermore aggravated by emotional stress and time constraints.

\begin{figure*} % Diagnostic Mode Overview
    \centering
    \includegraphics[width = 1.00\textwidth]{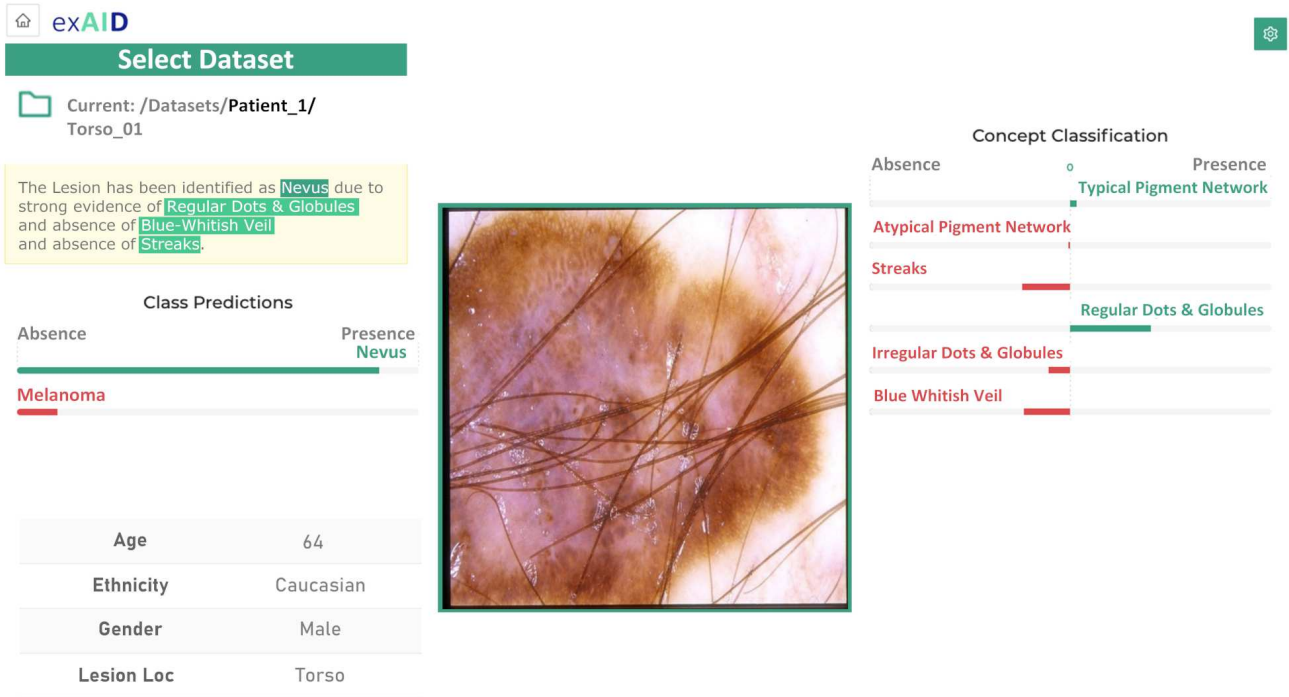}
    \caption{Diagnostic Mode of ExAID can be used as Decision Support System in routine clinical workflow.}
    \label{fig:modes_diagnostic}
\end{figure*}

ExAID's diagnostic mode aims at mediating subjectivity by offering a supplement to the experienced physician's first impressions, serving as a second opinion which stimulates physician's thought and breaks the routine.
Through this additional information it is made sure that cues vital for successful identification of malignant conditions are not overseen during manual examination.
The user interface of the diagnostic mode is presented in Fig~\ref{fig:modes_diagnostic}. 
While allowing physicians to examine the dermoscopic image manually, an initial diagnosis suggestion is provided, supported by concept-based textual, quantitative and visual explanations. 
Through its neutral design, the interface assures that users are not biased towards the proposed diagnosis but are free to reconstruct the AI's decision process by considering and validating biomarker scores along with their optional localisations provided in the form of CLMs. 

\begin{figure*} % Edu Mode Overview
    \centering
    \includegraphics[width = 1.0\textwidth]{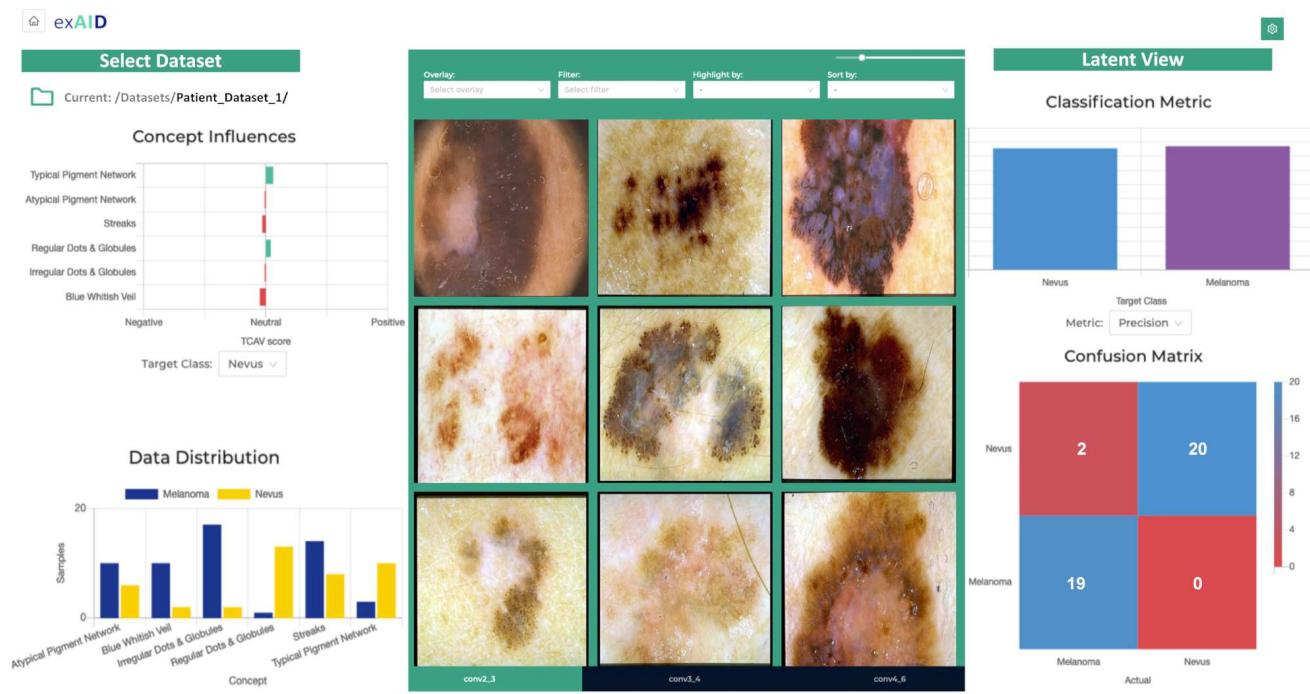}
    \caption{Educational Mode of ExAID can help in training of resident dermatologists by allowing them to explore many of its interactive features.}
    \label{fig:modes_educational}
\end{figure*}

\subsubsection{Educational Mode}
Explanation of classifiers' decisions has further utility beyond mere information and guidance of the algorithm's users. 
It is of central importance for the validation of individual automated decisions, verification of plausibility of a model's global generalisation behaviour and can additionally aid the decryption of unintelligible, decision relevant concepts learned by the AI. 
With its educational mode, as presented in Fig.~\ref{fig:modes_educational}, ExAID offers an extensive toolbox for the investigation of model behaviour and data distribution. 
Dataset-level model behaviour analysis is enabled through a combination of class-wise performance evaluation metrics and concept-wise global explanation metrics in combination with tools for facilitated overview of individual decision outcomes and explanations.
Some of the most salient interactive features of ExAID framework are introduced below.
\paragraph{Filtering}
The filtering option allows to filter arbitrary subsets of samples by metadata such as age, concept presence, concept prediction or correct prediction. 
An adaptive data distribution plot helps to quickly identify important statistical characteristics related to biomarker presence as well as certain failure modes of the model.
\paragraph{Highlighting}
A highlighting feature allows to spotlight certain useful properties of samples to further facilitate review of model behaviour and data. 
This feature allows the highlighting of not only binary attributes such as the correct target class prediction, but also more complex relationships such as the presence of classes or concepts in the annotations as well as the class and concept prediction by the model. 
Complex highlighting is always supported by visual cues indicating the accordance of attribute prediction with expert annotations.
\paragraph{Localisation}
In addition to individual localisation of concepts in data samples, ExAID allows to visualise concept localisation simultaneously for all samples of a dataset.
This allows for quick examination of a model's concept localisation behaviour, aiding the validation of system behaviour and identification of potential systematic errors in dataset or model by revealing patterns in the localisation process.
\paragraph{Latent Inspection}
Examination of the model's latent space structure gives further insight into the disentanglement of data representations and potential biases captured by the model parameters.
A latent view functionality based on Tensorboard's projector\footnote{https://projector.tensorflow.org/} allows to intuitively examine the latent distribution of data samples by means of dimensionality reduction techniques.

%%%%%%%%%%%%%%%%%%%%%%%%%%%%%%%% Results of Image Classification %%%%%%%%%%%%%%%%%%%%%%%%%%%%%%%%%%%%
\begin{table}[b!]
\centering
\caption{Performance evaluation of lesion classifier on various datasets.}
\label{tab:resultsImgClf}
\scalebox{0.9}{
\begin{tabular}{@{}lccccc@{}}
\toprule
\multirow{2}{*}{\textbf{Datasets}} &
\multirow{2}{*}{\textbf{N}} &
\textbf{Accuracy} &
\textbf{Precision} &
\textbf{Recall} &
\multirow{2}{*}{\textbf{AUC}} \\ 
 & & (\%) & (\%) & (\%) & \\ \midrule
Derm7pt (Test)          & 165 & 83.6                                             & 81.7 & 78.0 & 0.85 \\
PH2 (Test)              & 40 & 100.0                                            & 100.0 & 100.0 & 1.00 \\
ISIC2019 (Test)         & 1295 & 88.9                                             & 88.2 & 84.9 & 0.91 \\
ISIC2017 (Test)         & 510 & 78.4 & 68.5 & 62.3 & 0.70 \\
ISIC2016 (Test)         & 379 & 89.7                                             & 83.7 & 84.0 & 0.92 \\
SkinL2                  & 55 & 90.9                                             & 89.9 & 90.7 & 0.99 \\ \bottomrule
\end{tabular}
}
\end{table}

%%%%%%%%%%%%%%%%%%%%%%%%%%%%%%%%%%% Experiments and Results %%%%%%%%%%%%%%%%%%%%%%%%%%%%%%%%%%%%

\section{Results}
\label{sec:experiments_and_results}

\subsection{Classifier Training \& Evaluation}
To demonstrate the utility of the proposed framework, a deep network for binary classification of \textit{Melanoma} and \textit{Nevi} from dermoscopic skin lesion images is trained. 
Among various architecture, learning rate and optimiser combinations\footnote{Experimentation included VGG16, ResNet, DenseNet, NASNet, SEResNeXt architectures with Adam, SGD and RMSprop optimisers using learning rates ranging from 1e-3 to 1e-4}, best results have been achieved using SEResNeXt architecture with RMSprop optimiser and a learning rate of 1e-4 trained for 100 epochs. 
Training images were augmented by random horizontal and vertical flip as well as random cropping to 85\% of the image size, resulting in input images of size $224 \times 224$. 

Evaluation on a variety of datasets is presented in Table~\ref{tab:resultsImgClf}.
It can be observed that the lesion classifier achieved AUCs above 0.85 for five out of six datasets with two datasets scoring almost perfectly.
ISIC2017 achieved a slightly lower AUC with 0.70, which is also reflected in lower Precision and Recall.

%%%%%%%%%%%%%%%%%%%%%%%%%%%%%%%% Results of Concept Classification %%%%%%%%%%%%%%%%%%%%%%%%%%%%%%%%%%%%
\begin{table}
\centering
\caption{Performance evaluation of concept classifiers on various datasets. Results are given as Macro Average F1-Scores to account for class imbalance.}
\label{tab:resultsConceptClf}
\scalebox{0.8}{
\begin{tabular}{@{}lccccccccc@{}}
\toprule
\multirow{2}{*}{\textbf{Datasets}} &
\multirow{2}{*}{\textbf{Streaks}} &
\textbf{Pigment} &
\textbf{Dots \&} &
\textbf{Regr.} &
\textbf{Blue-Whit.} \\ 
 & & \textbf{Netw.} & \textbf{Glob.} & \textbf{Struct.} & \textbf{Veils} \\ \midrule
D7PH2 (Test)            & 70.91 & 78.74 & 63.14 & 59.41 & 71.66 \\
ISIC2017                & 51.75 & 50.37 &   -   &   -   &   -   \\
ISIC2016                & 56.53 &   -   & 53.03 &   -   &   -   \\ \bottomrule
\end{tabular}
}
\end{table}

\subsection{Explanation Training \& Evaluation}
For the explanation of the final DL-based classifier's decisions, the procedure outlined in~\cite{lucieri2020interpretability} is followed.
To this end, concept annotated samples from D7PH2 have been utilised to assure generalisation while learning CAVs. 
In each run, the data is internally split into folds for concept training and validation under stratification of both concept and disease labels.
For each concept, linear concept classifiers are trained for 200 runs using stochastic gradient descent with early stopping.

\subsubsection{Concept Detection}
The final CAV for a concept is chosen based on the average concept direction over all runs.
Due to concept annotation requirements, concept detection performance is evaluated only on ISIC2016 and ISIC2017 datasets as well as D7PH2 test set.
Due to the lack of annotation, the two ISIC datasets allowed evaluation of only two concepts each.
Table~\ref{tab:resultsConceptClf} presents Macro Average F1-Scores for concept detection.

For D7PH2 test set, all concept detectors were able to discriminate concepts better than random guessing, with \textit{Pigment Network} achieving the best F1-Score of 78.74\%.
Same holds for the evaluation on ISIC2016 for concepts \textit{Streaks} and \textit{Dots \& Globules}.
Similar to the results for lesion classification on ISIC2017, concept detectors failed to classify \textit{Streaks} and \textit{Pigment Networks}, yielding F1-Scores of 51.75\% and 50.37\%, respectively.

%%%%%%%%%%%%%%%%%%%%%%%%%%%%%%%%%% Figure: CLM Examples %%%%%%%%%%%%%%%%%%%%%%%%%%%%%
\begin{figure*}
    \centering
    \setkeys{Gin}{width=\linewidth} %set image parameters
    \begin{subfigure}{8cm}
    \centering
        \includegraphics{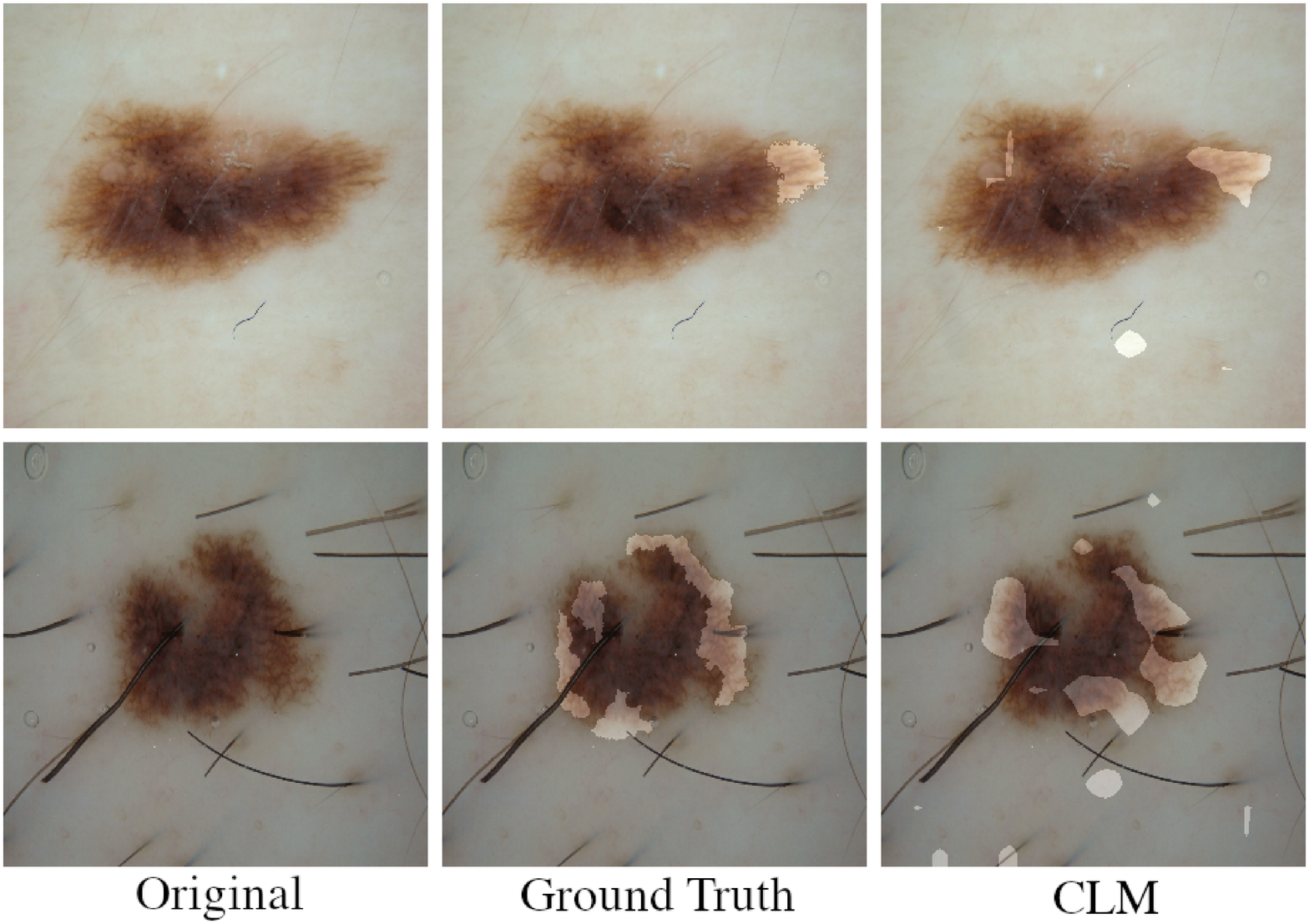}
        \caption{Streaks.}
        \label{fig:CLMpos_ST}
    \end{subfigure}
    \hfil
    \begin{subfigure}{8cm}
        \centering
        \includegraphics{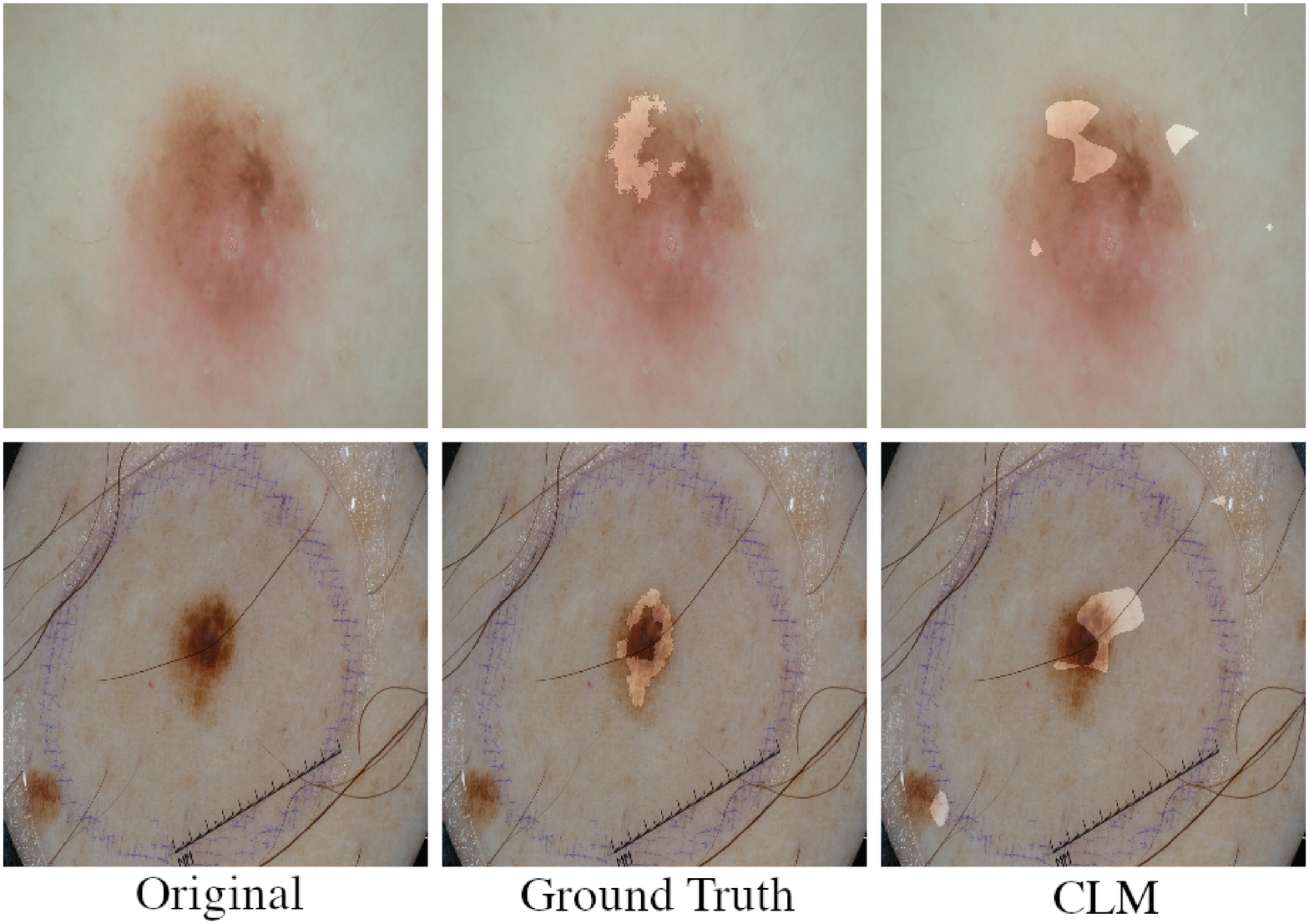}
        \caption{Pigment Network.}
        \label{fig:CLMpos_PN}
    \end{subfigure}
    \caption{Positive and negative examples of visual explanations provided by ExAID along with the corresponding samples and ground truth concept masks.}
    \label{fig:CLMpos}
\end{figure*}

\subsubsection{Visual Explanation}
Fair quantitative evaluation of a network's CLMs for skin lesions poses a number of difficulties including the selection of a suitable binarization scheme, subjectivity of concept annotations as well as lack of representative metrics for fuzzy localisation tasks.
Proper binarization is specially difficult as it depends on the size of a particular Region of Interest (ROI), its significance to the prediction score as well as further noise stemming from the saliency method used.
Moreover, evaluation is limited by the availability of annotated concept segmentation maps. 
ISIC2016 and ISIC2017 challenge datasets each provide concept segmentation maps for two concepts which are used to provide a qualitative assessment of the trained model's concept localisation ability.
CLMs were binarized using variable percentiles, manually chosen based on the size of the respective ROI in a specific image.
Figure~\ref{fig:CLMpos} shows examples of the model's concept localisation ability for classes \textit{Streaks} and \textit{Pigment Network} using an adaptation of the method proposed in~\cite{lucieri2020CLM}.
Interpretation of the results is provided in Section~\ref{sec:concept_localisation}.

\subsubsection{Textual Explanation}
\begin{figure*}
    \centering
    \setkeys{Gin}{width=\linewidth} %set image parameters
    \begin{subfigure}{8cm}
    \centering
        \includegraphics{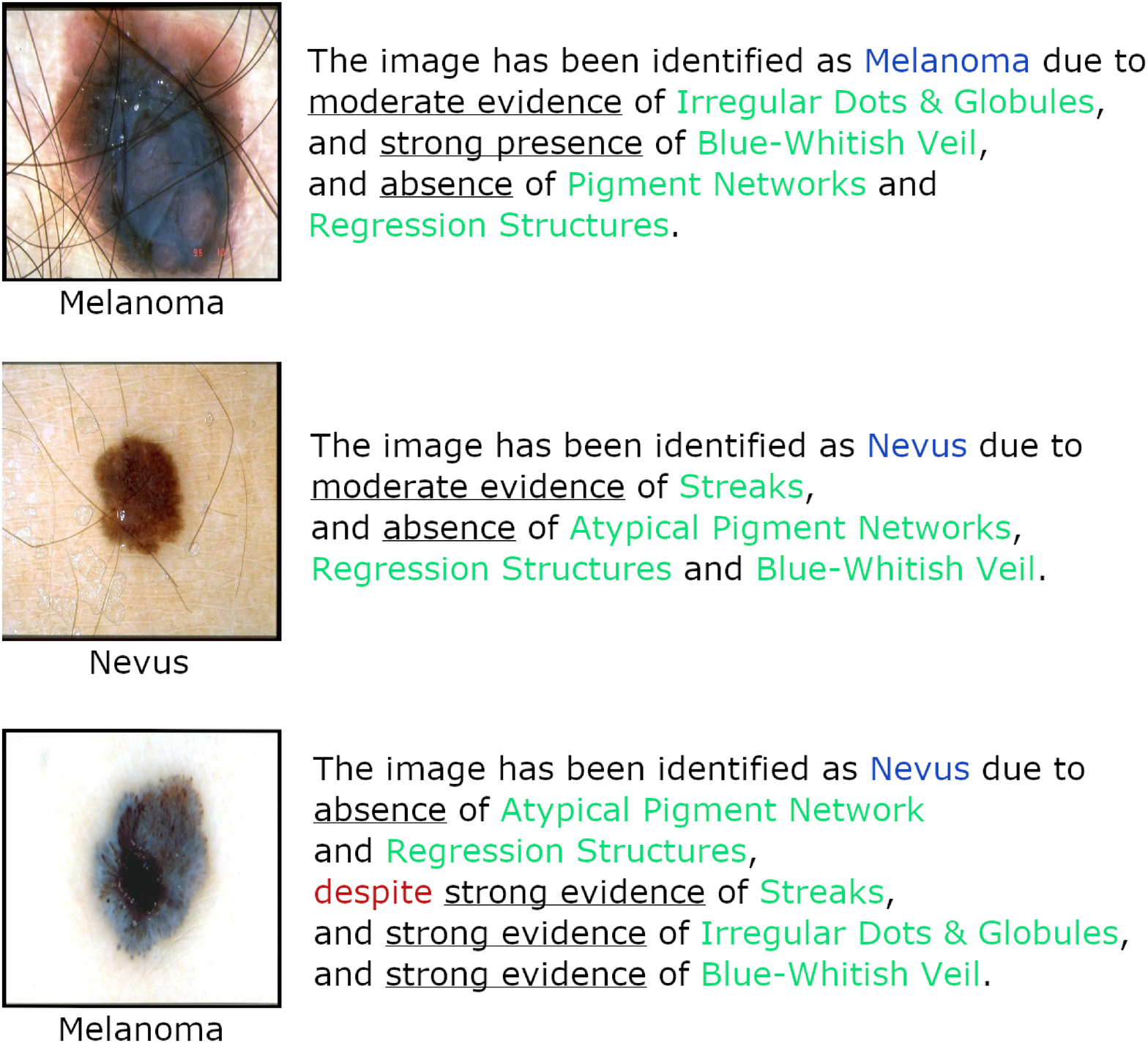}
        \caption{Correct concept prediction.}
        \label{fig:examples_verbal_pos}
    \end{subfigure}
    \hfil
    \begin{subfigure}{8cm}
        \centering
        \includegraphics{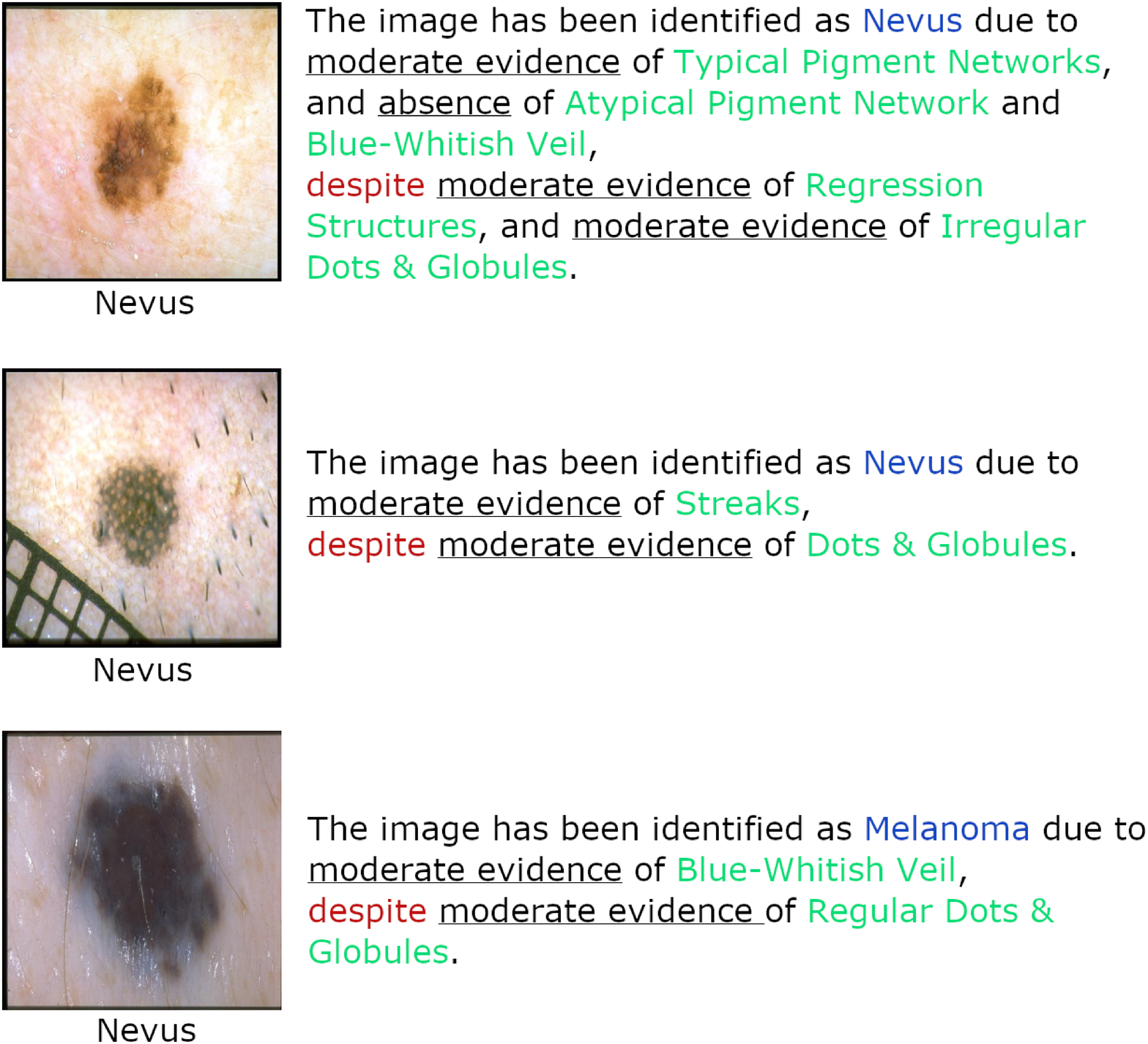}
        \caption{Incorrect concept prediction.}
        \label{fig:examples_verbal_neg}
    \end{subfigure}
    \caption{Positive and negative examples of textual explanations provided by ExAID along with the corresponding skin lesion samples. The ground truth class of the sample is given below the image.}
    \label{fig:examples_verbal}
\end{figure*}

Quantitative evaluation of textual explanation results is covered by the performance evaluation for concept detection presented in table~\ref{tab:resultsConceptClf}.
Figure~\ref{fig:examples_verbal} shows qualitative examples of images along with correct and incorrect textual explanations provided by ExAID.
These results are further discussed in Section~\ref{sec:textual_explanation}.
    
%%%%%%%%%%%%%%%%%%%%%%%%%%%%%%%%%% Figure: Verbal Examples %%%%%%%%%%%%%%%%%%%%%%%%%%%%%

\section{Discussion}
\label{sec:discussion}
The results presented in Section~\ref{sec:experiments_and_results} show that ExAID is indeed able to produce theoretically correct explanations for a classifier's decisions.
In the following, previous insights are discussed and qualitative results are analysed in detail.
Finally, a detailed discussion about the limitation and future work is provided.

\subsection{Lesion Classification}
\label{sec:lesion_classification}
Lesion-level results clearly show the strong generalisation ability of the model, even on unseen datasets as SKINL2 consisting of 20 Melanomas and 35 Nevi of high quality.
Poor performance on the ISIC2017 test dataset can be explained by the large fraction of artefacts present in the images, which have been intentionally left out of the training procedure (through manual cleansing) to restrict the use case to a realistic, controlled diagnostic environment based on an image acquisition procedure specifically built for AI processing. 

\subsection{Concept Detection}
\label{sec:concept_detection}
In contrast to the concept detection performance on the D7PH2 test set, concept generalisation to unseen datasets such as ISIC2017 and ISIC2016 is worse.
This is most likely a consequence of diverging annotation standards between derm7pt and PH\textsuperscript{2} datasets used for CAV training and other datasets.
The distribution shift partially caused by artefacts present in the challenge test sets aggravates this divergence further. 
Moreover, results show the superiority of coarse-grained biomarkers such as \textit{Streaks}, \textit{Pigment Networks} and \textit{Blue-Whitish Veils} over more fine-grained ones such as \textit{Dots \& Globules}.

\subsection{Visual Explanation}
\label{sec:concept_localisation}
Whereas in some cases, CLM localisation aligned very well with the concept annotation, most of the time CLMs highlighted slightly different regions.
However, these highlights often depict areas that could plausibly count as concept regions, as can be seen in the second row of Figure~\ref{fig:CLMpos}.
Qualitative evaluation confirmed the quantitative results and showed that the network performed better localising concepts \textit{Streaks} and \textit{Pigment Networks} as compared to the more fine-grained \textit{Dots \& Globules} concept. 
Scattered spots in CLMs outside the lesion regions highlights noise problems inherent in perturbation-based CLM computation and the dependence on a proper binarization scheme.
%specially when evaluating maps quantitatively.

\subsection{Textual Explanation}
\label{sec:textual_explanation}
Examples in Figure~\ref{fig:examples_verbal_pos} depict instances with correct concept predictions, which showcase the simplicity and intelligibility of the generated explanations.
Explanation texts briefly reflect the most important criteria necessary for experts to understand the network's decision.
Interestingly, it appeared that although correct concept predictions where given, the network sometimes misclassified the underlying disease as seen in the third row of Figure~\ref{fig:examples_verbal_pos}.
This could be due to the presentation of an ambiguous borderline case or a result of wrong ground truth annotation for either lesion class or concepts.
However, the explanation explicitly exposes \textit{Streaks}, \textit{Irregular Dots \& Globules} and \textit{Blue-Whitish Veil} as contraindications for the prediction of Nevus.
In a clinical setting, such contraindication would raise the suspicion of a user, possibly initiating a more thorough review of the case.
This particularly emphasises the utility of such a system, as a correct explanation will allow physicians to scrutinise a given prediction, not solely relying on an automated, opaque categorical output value.

Figure~\ref{fig:examples_verbal_neg} on the contrary shows failure cases where the network confused different visual cues for concepts.
While \textit{Irregular Dots \& Globules} have been correctly detected in the top right image, the middle right image contains white blobs which might have been confused as \textit{Dots \& Globules} by the model.
The bottom right case shows a \textit{Blue Nevus} which has been confused by the network as a \textit{Melanoma} showing signs of \textit{Blue-Whitish Veil} although containing \textit{Regular Dots \& Globules}.
It is notable that samples with incorrect concept predictions already expose a certain uncertainty by exhibiting moderate concept detections more frequently as compared to the samples from Figure~\ref{fig:examples_verbal_pos} as well as their prevalence of contraindications.
This clearly shows that irrespective of the model used for prediction, ExAID is able to provide well-founded justifications which help to express model uncertainty, encouraging closer examination of rare and edge cases.
%%%%%%%%%%%%%%%%%%%%%%%%%%%%%%%%%%%%%%%%%% Limitations %%%%%%%%%%%%%%%%%%%%%%%%%%%%%%%%%%%%%%%%%%
\subsection{Limitations}
\label{sec:limitations}
This proof-of-principle study primarily focuses on the current state of the proposed framework with its comprehensible user interface, conveying textual, visual and conceptual explanations for trustworthy computer-aided decision support in medicine.
The development of the framework is an ongoing Co-Design process which holistically includes a wide variety of stakeholders with different needs to assure not only practical but also ethical aspects early on~\cite{zicari2021co}.
Although, concept classification, localisation and textual explanation abilities of ExAID are remarkable given the fact that the DL model has not explicitly been trained on those tasks, some challenges must be first solved before an application in real clinical settings becomes feasible.

Current public datasets often suffer from the low sample quality attributable to a lack of process standardisation\footnote{Different camera setups, operators and techniques like polarised and non-polarised dermoscopy resulting in varying image quality, lighting, alignment and artefacts.}, missing histological diagnosis confirmation and subjective annotation.
Together with the low number of overall available images, in particular the ones with detailed concept annotation, this results in significant shift of data distributions between different datasets, constituting the major reason for sub-optimal generalisation of the proposed concept classifiers to other datasets.

ExAID's concept localisation ability yet suffers from limitations due to the perturbation-based nature of saliency map generation which results in noisy heatmaps and high sensitivity to hyperparameters, specially in case of varying sizes of biomarkers.
Future work applying optimisation-based perturbation methods for concept localisation will mitigate those issues, resulting in more flexible and robust heatmaps.
Textual explanations are generated based on concept predictions as well as directional derivatives as used in TCAV scores. 
Lacking a meaningful scaling of gradients, only the direction and not the magnitude of a concept's influence is currently used to improve the explanation text.
Incorporation of more robust concept influence measures could add another level of details to the rule-base, making the explanations more differentiated and rendering the system even more useful in practice.

Quantitative evaluation of concept detection or localisation is still limited due to lack of similarly and sufficiently annotated data from other sources.
To solve this issue, an agreed upon definition and consensual annotation of a large number of representative images is required, which will reflect in higher quality explanations.
Moreover, evaluation of CLMs is aggravated by noise artefacts emerging during binarisation and lack of definite measures for fuzzy localisation tasks.
A qualitative evaluation in real-world setting by medical experts is of extreme value for the evaluation of the explanations' utility to the diagnostic workflow and will be realised as soon as may be.

The influence of subjectivity not only reflects in the data annotations, but also in the general uncertainty surrounding the field of dermoscopy.
Despite first attempts towards standardisation of dermoscopic terminologies and concepts~\cite{kittler2016standardization}, no general consensus has yet been broadly established among physicians. 
Thus, a variety of diagnostic schools prevail and interpretation of terms and concepts is still largely depending on the education, preference and experience of the individual physician.
This work focuses on the 7-point checklist criteria~\cite{argenziano1998epiluminescence} as well as further dermoscopic concepts from~\cite{mendoncca2013ph}, due to the public availability of annotated data.
The commitment to a specific set of concepts prior to the decision of a standard consensus might hamper the acceptance of the framework by physicians accustomed to different methods and the mixture of different schools and interpretations of concepts bears the risk of contrasting labelling.
Productive deployment of such a system requires diligent assessment through medical practitioners in real-world environments, providing their valuable feedback to evaluate and improve such a system.
Prior to performing clinical trials, the system should be fed with carefully selected data properly representing a set of meaningful and unambiguously defined dermoscopic concepts as agreed by a committee of dermatology experts.

%We report only positive concept appearance but not absence as indicators in the explanation

%%%%%%%%%%%%%%%%%%%%%%%%%%%%%%%%%%%%%%%%%% Conclusion %%%%%%%%%%%%%%%%%%%%%%%%%%%%%%%%%%%%%%%%%%

\section{Conclusion}
\label{sec:conclusion}
Since the advent of modern DL-based systems and their industrious applications in medical domains, there have been remarkable strides in the explanation of these complex systems that, in some cases, already led to correction and verification of AI as well as disclosure of new potential diagnostic criteria. 
However, sensible and comprehensible explanations is still one of the greatest challenges related to medical image diagnosis, which should be addressed by concerted efforts from AI researchers, medical practitioners and regulatory authorities.

With ExAID, this article presents a framework which consolidates and builds upon our previous works on detection of human-defined concepts for skin lesion diagnosis in DL model's latent space and their localisation on the input image to provide intelligible textual explanation of model's predictions. 
We showed that, despite severe limitation in terms of data and annotation availability, the system already provides useful insights into DL classifier's decision making, even in case of wrong predictions.
When properly addressing the current limitations, this framework will not only play a useful assistive role in reliable, efficient and objective screening of melanoma, which is one of the most serious skin cancers, but also help train new dermatologists efficiently and effectively.

The generality of the framework allows its adaptation to various other image-based domains like radiology or histology.
In view of future deployment of the system in clinical practice, its under-lying general purpose DL architecture will be further specialised to the respective domain, including improvements to the processing pipeline such as lesion segmentation and hair removal.

\section*{Acknowledgement}
This work is partially funded by National University of Science and Technology (NUST), Pakistan through Prime Minister's Programme for Development of PhDs in Science and Technology and BMBF projects ExplAINN (01IS19074) and DeFuseNN (01IW17002).

\bibliographystyle{cas-model2-names}

\bibliography{cas-refs}
\clearpage

\bio{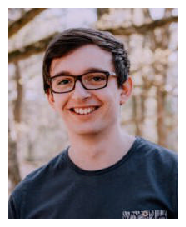}
\textbf{Adriano Lucieri} completed his BE in Mechatronic Engineering from Duale Hochschule Baden-Württemberg (DHBW) Mannheim and MS in Mechatronic Systems Engineering from Hochschule Pforzheim in Germany. He is presently pursuing PhD from Technische Universität Kaiserslautern (TUK), Germany and is also working as Research Assistant at Deutsches Forschungszentrum für Künstliche Intelligenz GmbH (DFKI). His research focus lies on improving the explainability and transparency of Computer-Aided Diagnosis (CAD) systems based on Deep Learning for medical image analysis. His work includes concept-based explanation of skin lesion classifiers as well as the localisation of concept regions in input images. 
\endbio

\bio{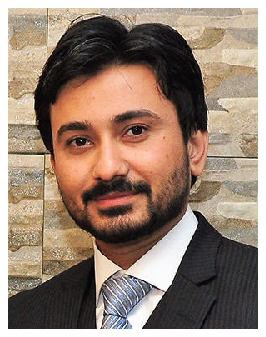}
\textbf{Muhammad Naseer Bajwa} completed his BS in Computer Engineering from COMSATS Institute of Information Technology (CIIT), Pakistan and MS in Computer Engineering from King Fahd University of Petroleum and Minerals (KFUPM), Saudi Arabia. He is presently pursuing PhD from Technische Universität Kaiserslautern (TUK), Germany and is also working as Research Assistant at Deutschen Forschungszentrums für Künstliche Intelligenz GmbH (DFKI). His main area of research is towards realising a practically usable, confident and interpretable Computer-Aided Diagnosis (CAD) system. He has published his works in various peer reviewed journals and top ranked conferences on detection of ocular disorders like glaucoma and diabetic retinopathy using retinal fundus images, automated diagnosis of cutaneous diseases using dermoscopic images, curation of retinal fundus images dataset for glaucoma detection and segmentation of optic disc and cup (G1020), and interpretability of CAD for skin lesions.
\endbio

\bio{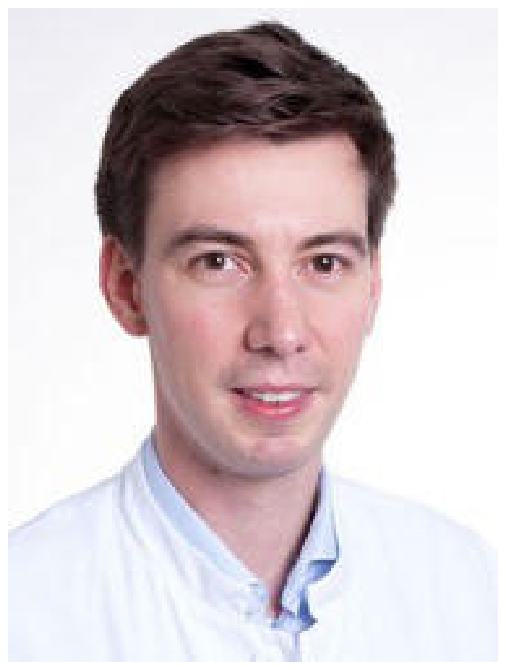}
\textbf{Stephan Alexander Braun} is a board-certified dermatologist and dermatopathologist and works at the University Hospital of Münster and Düsseldorf, Germany. His scientific work focuses on the diagnosis and treatment of skin tumors. \\ \\ \\ \\
\endbio

\bio{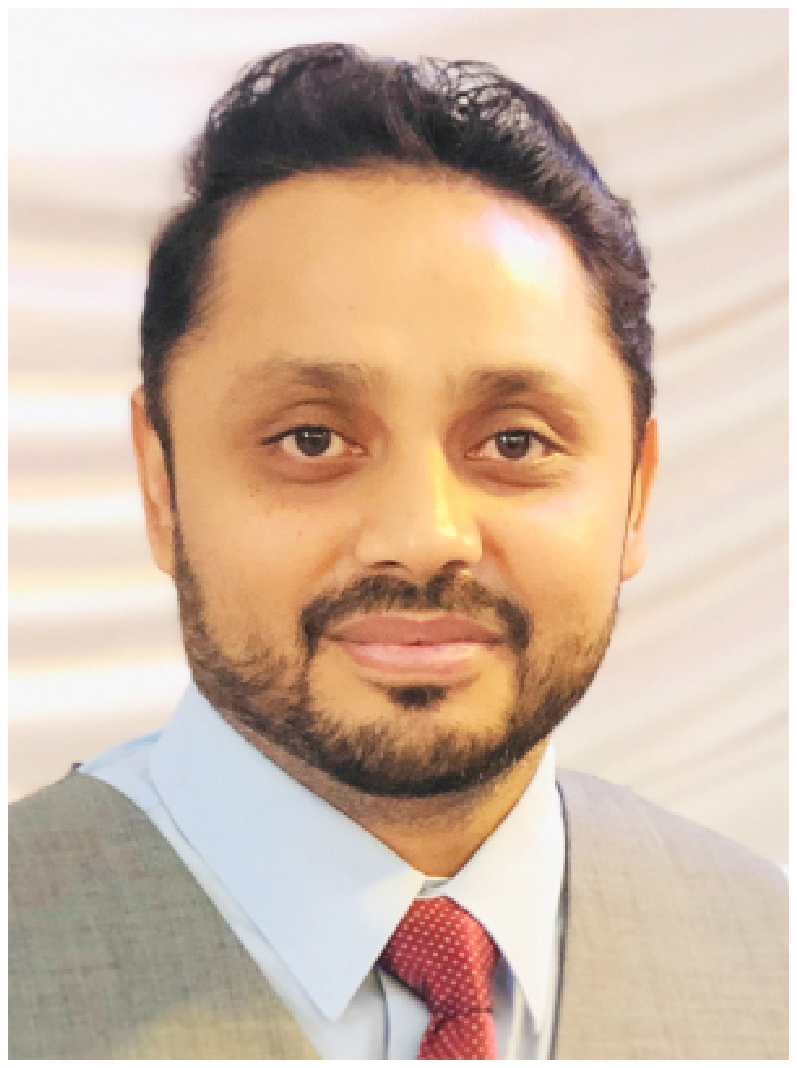}
\textbf{Muhammad Imran Malik} received  his  master's  and  PhD  degrees  in  Artificial  Intelligence,  in  2011  and  2015  respectively, from  the  University  of Kaiserslautern. He also worked in the German Research Center for Artificial Intelligence GmbH (DFKI), Kaiserslautern, Germany. His Ph.D. topic was automated forensic handwriting analysis on which he focused on both the perspectives of forensic handwriting examiners and pattern recognition researchers. He is currently an Assistant Professor with the School of Electrical Engineering and Computer Science (SEECS)  at  the  National  University  of Sciences  and  Technology (NUST), Islamabad, Pakistan. He has authored more than 40 publications including several journal and high ranked conference papers.
\endbio

\bio{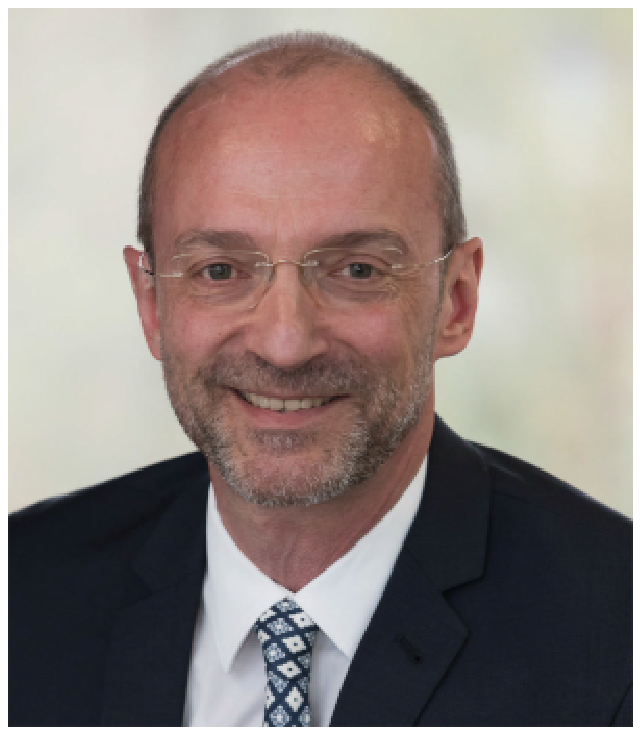}
\textbf{Andreas Dengel} is Scientific Director at DFKI GmbH in Kaiserslautern. In 1993, he became Professor in Computer Science at TUK where he holds the chair Knowledge-Based Systems. Since 2009 he is appointed Professor (Kyakuin) in Department of Computer Science and Information Systems at Osaka Prefecture University. He received his Diploma in CS from TUK and his PhD from University of Stuttgart. He also worked at IBM, Siemens, and Xerox Parc. Andreas is member of several international advisory boards, has chaired major international conferences, and founded several successful start-up companies. He is co-editor of international computer science journals and has written or edited 12 books. He is author of more than 300 peer-reviewed scientific publications and supervised more than 170 PhD and master theses. Andreas is an IAPR Fellow and received many prominent international awards. His main scientific emphasis is in the areas of Pattern Recognition, Document Understanding, Information Retrieval, Multimedia Mining, Semantic Technologies, and Social Media.
\endbio

\bio{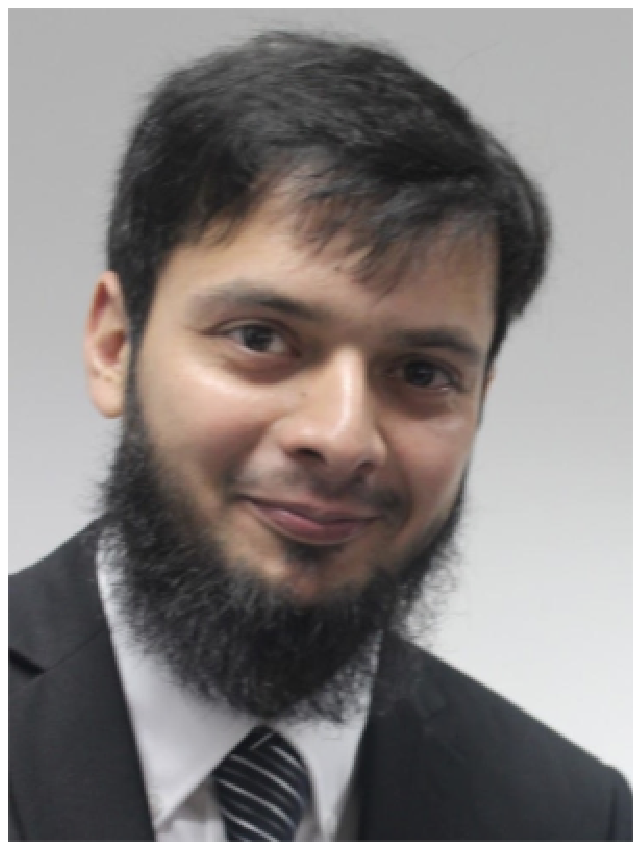}
\textbf{Sheraz Ahmed} is Senior Researcher at DFKI GmbH in Kaiserslautern, where he is leading the area of Time Series Analysis. He received his MS and PhD degrees in Computer Science from TUK, Germany under the supervision of Prof. Dr. Prof. h.c. Andreas Dengel and Prof. Dr. habil. Marcus Liwicki. His PhD topic is Generic Methods for Information Segmentation in Document Images. Over the last few years, he has primarily worked on development of various systems for information segmentation in document images. His research interests include document understanding, generic segmentation framework for documents, gesture recognition, pattern recognition, data mining, anomaly detection, and natural language processing. He has more than 30 publications on the said and related topics including three journal papers and two book chapters. He is a frequent reviewer of various journals and conferences including Patter Recognition Letters, Neural Computing and Applications, IJDAR, ICDAR, ICFHR, and DAS.
\endbio

\end{document}